%% file: motionest.tex
\documentclass[10pt,twocolumn,letterpaper]{article}
\pdfoutput=1

\usepackage{cite}
\usepackage{cvpr}
\usepackage{times}
\usepackage{epsfig}
\usepackage{graphicx}
\usepackage{amsmath}
\usepackage{amssymb}
\usepackage{multirow}
\graphicspath{{figs/}}

\usepackage[pagebackref=true,breaklinks=true,letterpaper=true,colorlinks,bookmarks=false]{hyperref}

\cvprfinalcopy

\begin{document}
\input{macros.tex}

\title{Learning Motion Patterns in Videos}

\author{Pavel Tokmakov
\and
Karteek Alahari\vspace{0.3cm}\\
\large{Inria\thanks{Thoth team, Inria, Laboratoire Jean Kuntzmann, Grenoble, France.}\vspace{-0.2cm}}
\and
Cordelia Schmid\\
}
\maketitle

\begin{abstract}
\vspace{-0.3cm}
The problem of determining whether an object is in motion, irrespective of
camera motion, is far from being solved. We address this challenging task by
learning motion patterns in videos. The core of our approach is a fully
convolutional network, which is learned entirely from synthetic video sequences,
and their ground-truth optical flow and motion segmentation. This
encoder-decoder style architecture first learns a coarse representation of the
optical flow field features, and then refines it iteratively to produce motion
labels at the original high-resolution. We further improve this labeling with
an objectness map and a conditional random field, to account for errors in
optical flow, and also to focus on moving ``things'' rather than ``stuff''. The
output label of each pixel denotes whether it has undergone independent motion,
i.e., irrespective of camera motion. We demonstrate the benefits of this
learning framework on the moving object segmentation task, where the goal is to
segment all objects in motion. Our approach outperforms the top method on the
recently released DAVIS benchmark dataset, comprising real-world sequences, by
5.6\%. We also evaluate on the Berkeley motion segmentation database, achieving
state-of-the-art results.
\end{abstract}

\vspace{-0.45cm}
\section{Introduction}
\begin{figure}[t]
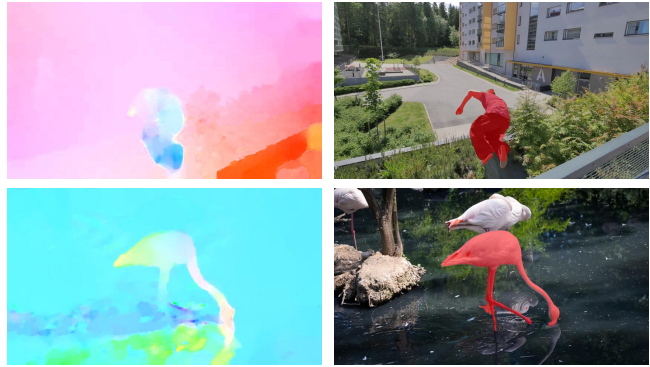

\begin{center}
\twofigures{intro/flow1.jpg}{intro/intro_ours1.jpg}{0.5\linewidth}\vspace{0.1cm}
\twofigures{intro/flow2.jpg}{intro/intro_ours2.jpg}{0.5\linewidth}
\end{center}
\vspace{-0.8cm}\caption{Results on the DAVIS dataset. Left: Optical flow field
input to our MP-Net, computed with~\cite{Brox11a}. Right: Our segmentation
result overlaid on the video frame. Note that our approach accurately segments
moving objects, and learns to distinguish between object and camera motions
(seen in the flow fields).\vspace{-0.6cm}}
\label{fig:result}
\end{figure}

\vspace{-0.2cm}
The task of analyzing motion patterns has a long history in computer
vision~\cite{Torr98,Bideau16,Dosovitskiy15,Horn81,Vedula05,papazoglou2013fast,XuC16}.
This includes methods for motion estimation~\cite{Torr98,Bideau16},
scene~\cite{Vedula05} and optical~\cite{Dosovitskiy15,Horn81} flow computation,
video segmentation~\cite{papazoglou2013fast,XuC16}; all of which aim to
estimate or capitalize on motion cues in scenes. Despite this progress, the
fundamental problem of identifying if an object is indeed moving, irrespective
of camera motion, remains challenging. In this paper, we make significant
advances to address this challenge, with a novel CNN-based framework to
automatically learn motion patterns in videos, and use it to segment moving
objects; see sample results in Figure~\ref{fig:result}.

\begin{figure*}[t]
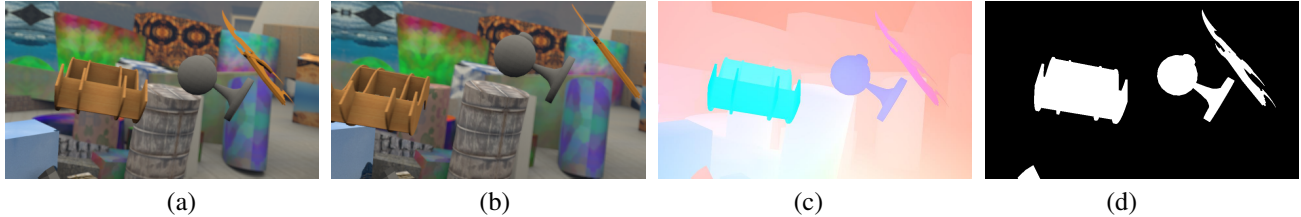

\begin{center}
\fourfigurescaption{2/exfrm1}{2/exfrm2}{2/exflow}{2/exgt}{0.5\columnwidth}{(a)}{(b)}{(c)}{(d)}
\end{center}
\vspace{-0.3cm}\caption{(a,b) Two example frames from a sequence in the
FlyingThings3D dataset~\cite{Mayer16}. The camera is in motion in this scene,
along with four independently moving objects. (c) Ground-truth optical flow of
(a), which illustrates motion of both foreground objects and background with
respect to the next frame (b). (d) Ground-truth segmentation of moving objects
in this scene. \vspace{-0.5cm}}
\label{fig:intro}
\end{figure*}

To illustrate the task, consider Figure~\ref{fig:intro}, a sequence from the
FlyingThings3D dataset~\cite{Mayer16}. It depicts a scene generated
synthetically, involving a moving camera (can be easily observed by comparing
the top left corners of the images (a) and (b)), with objects in motion, e.g.,
the three large objects in the centre of the frame (which are easier to spot in
the ground-truth segmentation (d)). The goal of our work is to study such
motion patterns in video sequences (using optical flow field (c)), and to learn
to distinguish real motion of objects from camera motion. In other words, we
target the moving object segmentation in (d).

The core of our approach is a trainable model, motion pattern network (MP-Net),
for separating independent object and camera motion, which takes optical flow
as input and outputs a per-pixel score for moving objects. Inspired by fully
convolutional networks (FCNs)~\cite{long2015fully,Dosovitskiy15,Ronneberger15},
we propose a related encoder-decoder style architecture to accomplish this
two-label classification task. The network is trained from scratch with
synthetic data~\cite{Mayer16}.  Pixel-level ground-truth labels for training
are generated automatically (see Figure~\ref{fig:intro}(d)), and denote whether
each pixel has moved in the scene. The input to the network is flow fields,
such as the one shown in Figure~\ref{fig:intro}(c). More details of the
network, and how it is trained are provided in Section~\ref{sec:model}. With
this training, our model learns to distinguish motion patterns of objects and
background. We then refine these labels with objectness
cues~\cite{pinheiro2016learning} and a conditional random field (CRF)
model~\cite{krahenbuhl2011efficient} (see \S\ref{sec:realvid}), to demonstrate
the efficacy of the entire framework on the moving object segmentation task
(see \S\ref{sec:expts}). These refinement steps are important to account for
errors in flow fields, and also to target moving objects, instead of
\textit{stuff} such as moving water. We evaluate on the densely annotated video
segmentation (DAVIS)~\cite{Perazzi16} and the Freiburg/Berkeley motion
segmentation datasets (BMS-26,
FBMS)~\cite{Bideau16,brox2010object,Tron07,ochs2014segmentation}, all
comprising real-data sequences. We obtain state-of-the-art results on these
challenging datasets. In particular, we outperform previous video-level methods
by over 5.6\%, on the intersection over union score, on DAVIS, despite
operating only on the frame level. We have made the source code and the trained
models available
online.\footnote{\url{http://thoth.inrialpes.fr/research/mpnet}}

\vspace{-0.2cm}
\section{Related Work}
\vspace{-0.2cm}
Our work is related to the following tasks dealing with motion cues: motion and
scene flow estimation, and video object segmentation. We will review the most
relevant work on these topics, in addition to a review of related CNN
architectures in the remainder of this section.\\

\vspace{-0.2cm}
\noindent {\bf Motion estimation.} Early attempts for estimating motion have
focused on geometry-based approaches, such as~\cite{Torr98}, where the
potential set of motions is identified with RANSAC. Recent approaches have
relied on other cues to estimate moving object regions. For example, Papzouglou
and Ferrari~\cite{papazoglou2013fast} first extract motion boundaries by
measuring changes in optical flow field, and use it to estimate moving regions.
They also refine this initial estimate iteratively with appearance features.
This approach produces interesting results, but is limited by its heuristic
initialization. We show that incorporating our learning-based motion estimation
into it improves the results significantly (see Table~\ref{tbl:bms}).

Narayana~\etal~\cite{Narayana13} use optical flow orientations in a
probabilistic model to assign pixels with labels that are consistent with their
respective real-world motion. This approach assumes pure translational camera
motion, and is prone to errors when the object and camera motions are
consistent with each other. Bideau~\etal~\cite{Bideau16} presented an
alternative to this, where initial estimates of foreground and background
motion models are updated over time, with optical flow orientations of the new
frames. This initialization is also heuristic, and lacks a robust learning
framework.  While we also set out with the goal of finding objects in motion, our
solution to this problem is a novel learning-based method. Scene flow, i.e., 3D
motion field in a scene~\cite{Vedula05}, is another form of motion estimation,
but is computed with additional information, such as disparity values computed
from stereo images~\cite{Huguet07,Wedel11}, or estimated 3D scene
models~\cite{Vogel15}. None of these methods follows a CNN-based learning
approach, in contrast to our MP-Net. \\

\begin{figure*}[th]
\begin{center}
\includegraphics[width=2.0\columnwidth]{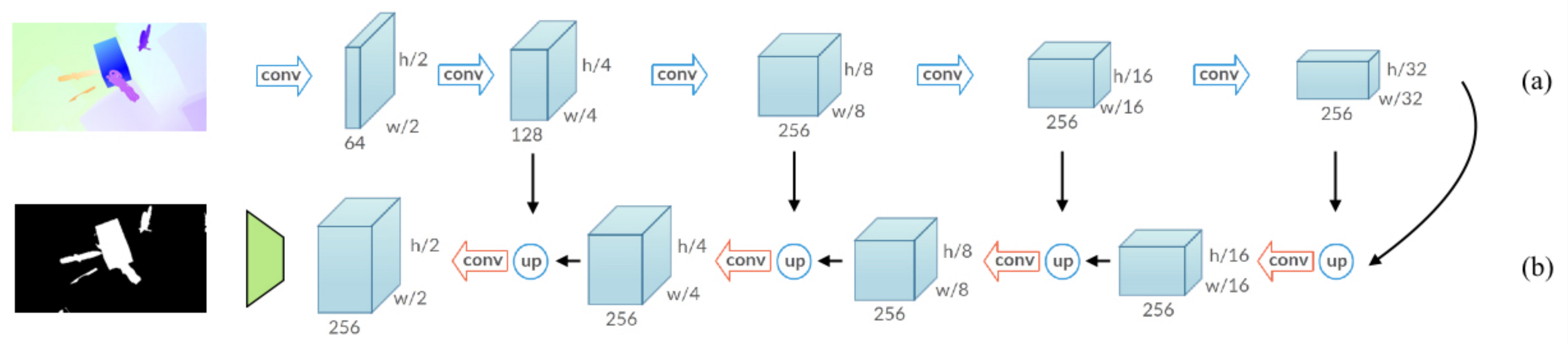}
\end{center}
\vspace{-0.4cm}
\caption{Our motion pattern network: MP-Net. The blue arrows in the encoder
part (a) denote convolutional layers, together with ReLU and max-pooling
layers. The red arrows in the decoder part (b) are convolutional layers with
ReLU, `up' denotes $2\times2$ upsampling of the output of the previous unit.
The unit shown in green represents bilinear interpolation of the output of the
last decoder unit. \vspace{-0.45cm}}
\label{fig:model}
\end{figure*}

\vspace{-0.2cm}
\noindent {\bf Video object segmentation.} The task of segmenting objects in video
is to associate pixels belonging to a class spatio-temporally; in other words,
extract segments that respect object boundaries, as well as associate object
pixels temporally whenever they appear in the video. This can be accomplished
by propagating manual segment labels in one or more frames to the rest of the
video sequence~\cite{Badrinarayanan10}. This class of methods is not applicable
to our scenario, where no manual segmentation is available.

Our approach to solve the segmentation problem does not require any
manually-marked regions. Several methods in this paradigm generate an
over-segmentation of
videos~\cite{Brendel09,Grundmann10,Lezama11,XuC16,Khoreva15}. While this can be
a useful intermediate step for some recognition tasks in video, it has no
notion of objects. Indeed, most of the extracted segments in this case do not
directly correspond to objects, making it non-trivial to obtain video object
segmentation from this intermediate result. An alternative to this is motion
segmentation~\cite{brox2010object,fragkiadaki2012video,ochs2012higher}, which
produces more coherent regions with point trajectories. They, however, assume
homogeneity of motion over the entire object, which is not valid for non-rigid
objects.

Another class of segmentation methods cast the problem as a
foreground-background classification
task~\cite{Faktor14,papazoglou2013fast,wang2015saliency,taylor2015causal,zhang2013video,lee2011key}.
Some of these first estimate a
region~\cite{papazoglou2013fast,wang2015saliency} or
regions~\cite{lee2011key,zhang2013video}, which potentially correspond(s) to
the foreground object, and then learn foreground/background appearance models.
The learned models are then integrated with other cues, e.g., saliency
maps~\cite{wang2015saliency}, pairwise
constraints~\cite{papazoglou2013fast,zhang2013video}, object shape
estimates~\cite{lee2011key}, to compute the final object segmentation.
Alternatives to this framework have used: (i) long-range interactions between
distinct parts of the video to overcome noisy initializations in low-quality
videos~\cite{Faktor14}, and (ii) occluder/occluded relations to obtain a
layered segmentation~\cite{taylor2015causal}. Our proposed method outperforms
all the top ones from this class of segmentation approaches (see
\S\ref{sec:expts}).\\

\vspace{-0.4cm}
\noindent {\bf Related CNN architectures.} Our CNN model predicts labels for every
pixel, similar to CNNs for other tasks, such as semantic
segmentation~\cite{long2015fully,Hariharan15,Ronneberger15}, optical
flow~\cite{Dosovitskiy15} and disparity/depth~\cite{Mayer16}
estimation. We adopt an encoder-decoder style network, inspired by the success
of similar architectures in~\cite{Dosovitskiy15,long2015fully,Ronneberger15}.
They first learn a coarse representation with receptive fields of gradually
increasing sizes, and then iteratively refine it with upconvolutional layers,
i.e., by upsampling the feature maps and performing convolutions, to obtain an
output at the original high-resolution. In contrast
to~\cite{Dosovitskiy15,long2015fully}, which predict labels in each of the
upconvolutional layers, we concatenate features computed at different
resolutions to form a strong representation, and estimate the labels in the
last layer. Our architecture also has fewer channels in the layers in the
encoding part, compared to~\cite{Ronneberger15}, to accommodate larger
training-set batches, and thus decrease the training time. More details of our
architecture are presented in Section~\ref{sec:net}.

\section{Learning Motion Patterns}
\label{sec:model}
\vspace{-0.05cm}
Our MP-Net takes the optical flow field corresponding to two consecutive frames
of a video sequence as input, and produces per-pixel motion labels. In other
words, we treat each video as a sequence of frame pairs, and compute the labels
independently for each pair. As shown in Figure~\ref{fig:model}, the network
comprises several ``encoding'' (convolutional and max-pooling) and ``decoding''
(upsampling and convolutional) layers. The motion labels are produced by the
last layer of the network, which are then rescaled to the original image
resolution (see \S\ref{sec:net}). We train the network entirely on synthetic
data---a scenario where ground-truth motion labels can be acquired easily (see
\S\ref{sec:train}).

\subsection{Network architecture}
\label{sec:net}
\vspace{-0.05cm}
Our encoder-decoder style network is motivated by the goal of segmenting
diverse motion patterns in flow fields, which requires a large receptive field
as well as an output at the original image resolution. A large receptive field
is critical to incorporate context into the model. For example, when the
spatial region of support (for performing convolution) provided by a small
receptive field falls entirely within an object with non-zero flow values, it
is impossible to determine whether it is due to object or camera motion. On the
other hand, a larger receptive field will include regions corresponding to the
object as well as background, providing sufficient context to determine what is
moving in the scene. The second requirement of output generated at the original
image resolution is to capture fine details of objects, e.g., when only a part
of the object is moving. Our network satisfies these two requirements with: (i)
the encoder part learning features with receptive fields of increasing sizes,
and (ii) the decoder part upsampling the intermediate layer outputs to finally
predict labels at the full resolution.

Figure~\ref{fig:model} illustrates our network architecture. Optical flow
field input is processed by the encoding part of the network (denoted by (a) in
the figure) to generate a coarse representation that is a $32\times32$
downsampled version of the input. Each 3D block here represents a feature map
produced by a set of layers. In the encoding part, each feature map is a result
of applying convolutions, followed by a ReLU non-linearity layer, and then a
$2\times2$ max-pooling layer. The coarse representation learned by the final set
of operations in this part, i.e., the $32\times32$ downsampled version, is
gradually upsampled by the decoder part ((b) in the figure). In each decoder
step, we first upsample the output of the previous step by $2\times2$, and
concatenate it with the corresponding intermediate encoded representation,
before max-pooling (illustrated with black arrows pointing down in the figure).
This upscaled feature map is then processed with two convolutional layers,
followed by non-linearities, to produce input for the next (higher-resolution)
decoding step. The final decoder step produces a motion label map at half the
original resolution. We perform a bilinear interpolation on this result to
estimate labels at the original resolution.

\begin{figure}[t]
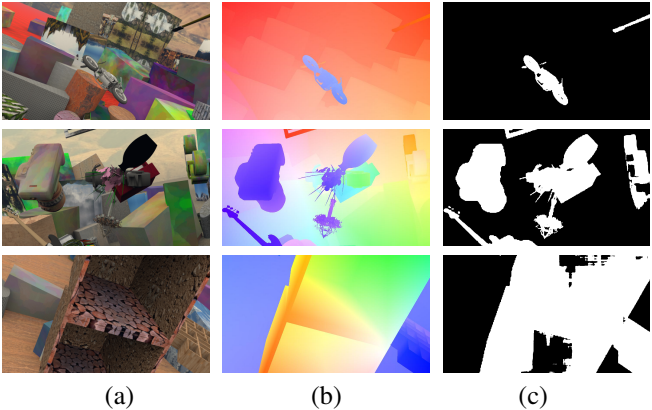

\begin{center}
\threefigures{6/exfrm1}{6/exflow}{6/exresult}{0.33\columnwidth} \vspace{0.1cm}
\threefigures{3/exfrm1}{3/exflow}{3/exresult}{0.33\columnwidth} \vspace{0.1cm}
\threefigurescaption{8/exfrm1}{8/exflow}{8/exresult}{0.33\columnwidth}{(a)}{(b)}{(c)}
\end{center}
\vspace{-0.3cm}\caption{Each row shows: (a) example frame from a sequence in
FlyingThings3D, (b) ground-truth optical flow of (a), which illustrates motion
of both foreground objects and background, with respect to the next frame, and
(c) our estimate of moving objects in this scene with ground-truth optical flow
as input.}
\vspace{-0.6cm}
\label{fig:qualff}
\end{figure}
\subsection{Training with synthetic data}
\label{sec:train}
We need a large number of fully-labelled examples to train a convolutional
network such as the one we propose. In our case, this data corresponds to
videos of several types of objects, captured under different conditions (e.g.,
moving or still camera), with their respective moving object annotations. No
large dataset of real-world scenes satisfying these requirements is currently
available, predominantly due to the cost of generating ground-truth annotations
and flow for every frame. We adopt the popular approach of using synthetic
datasets, followed in other work~\cite{Dosovitskiy15,Mayer16,Gaidon16}.
Specifically, we use the FlyingThings3D dataset~\cite{Mayer16} containing 2250
video sequences of several objects in motion, with ground-truth optical flow.
We augment this dataset with ground-truth moving object labels, which are
accurately estimated using the disparity values and camera parameters available
in the dataset, as outlined in Section~\ref{sec:datasets}. See
Figure~\ref{fig:intro}(d) for an illustration.

We train the network with mini-batch SGD under several settings. The one
trained with ground-truth optical flow as input shows the best performance.
This is analyzed in detail in Section~\ref{sec:mod}. Note that, while we use
ground-truth flow for training and evaluating the network on synthetic
datasets, all our results on real-world test data use only the estimated
optical flow. After convergence of the training procedure, we obtain a learned
model for motion patterns.

Our approach capitalizes on the recent success of CNNs for pixel-level
labelling tasks, such as semantic image segmentation, which learn feature
representations at multiple scales in the RGB space. The key to their top
performance is the ability to capture local patterns in images. Various types
of object and camera motions also produce consistent local patterns in the flow
field, which our model is able to learn to recognize. This gives us a clear
advantage over other pixel-level motion estimation
techniques~\cite{Bideau16,Narayana13} that can not detect local patterns.
Motion boundary based heuristics used in~\cite{papazoglou2013fast} can be seen
as one particular type of pattern, representing independent object motion. Our
model is able to learn many such patterns, which greatly improves the quality
and robustness of motion estimation.

\setlength{\tabcolsep}{2pt}
\begin{figure*}[t]
\begin{center}
\begin{tabular}{cccccc}
\includegraphics[width=0.33\columnwidth]{objectness/frame2} & 
\includegraphics[width=0.33\columnwidth]{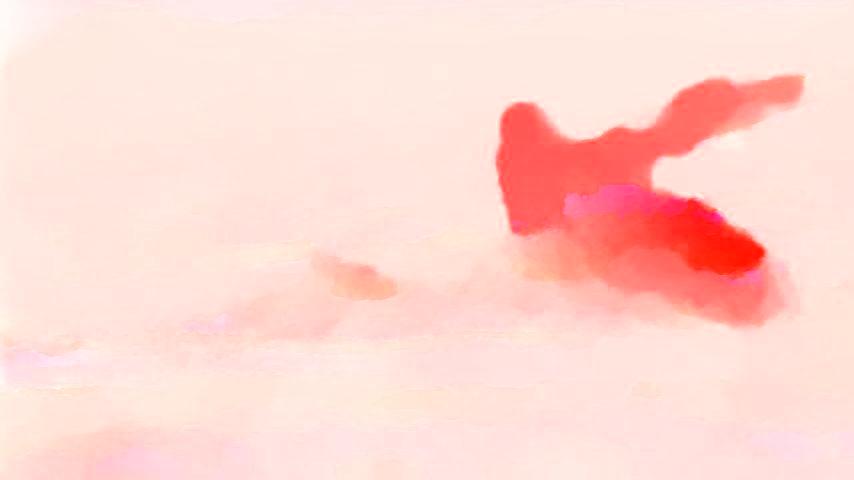} & 
\includegraphics[width=0.33\columnwidth]{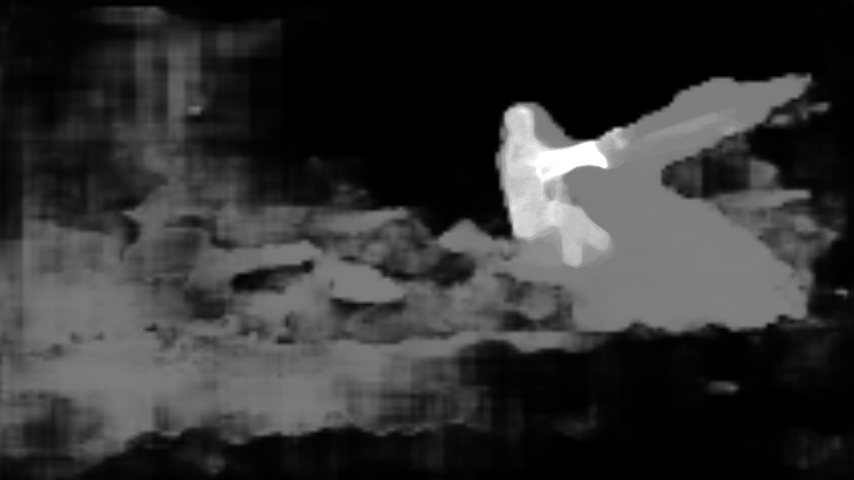} & 
\includegraphics[width=0.33\columnwidth]{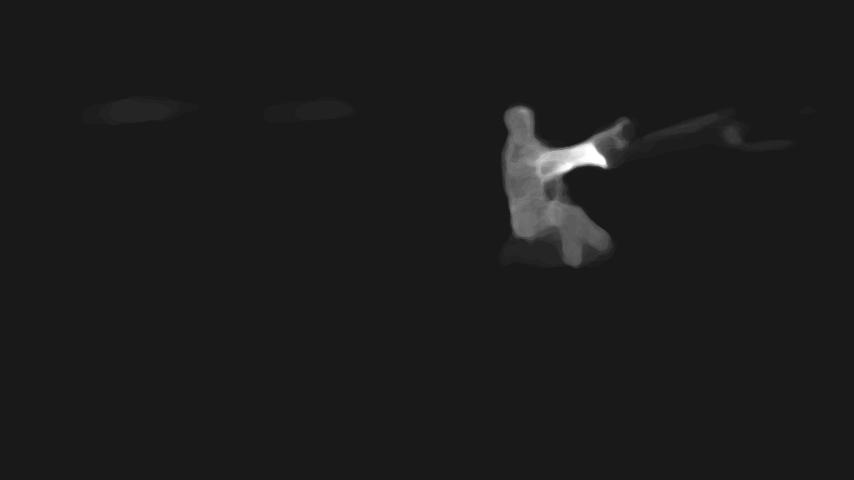} & 
\includegraphics[width=0.33\columnwidth]{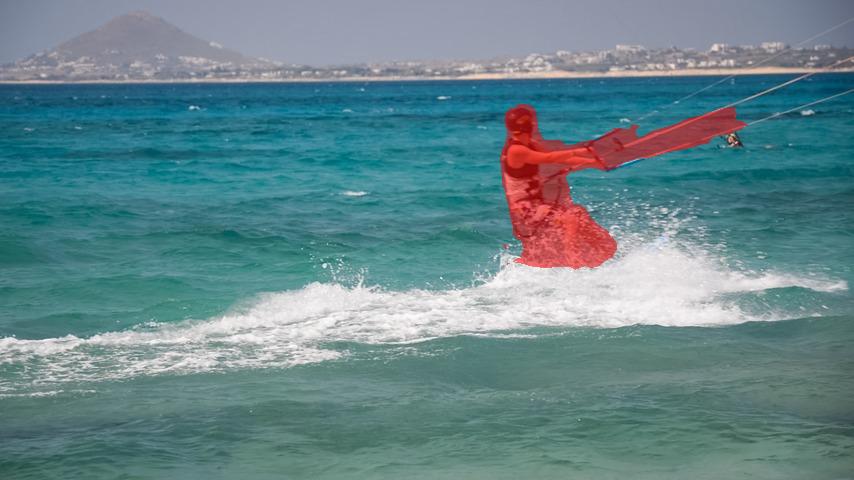} & 
\includegraphics[width=0.33\columnwidth]{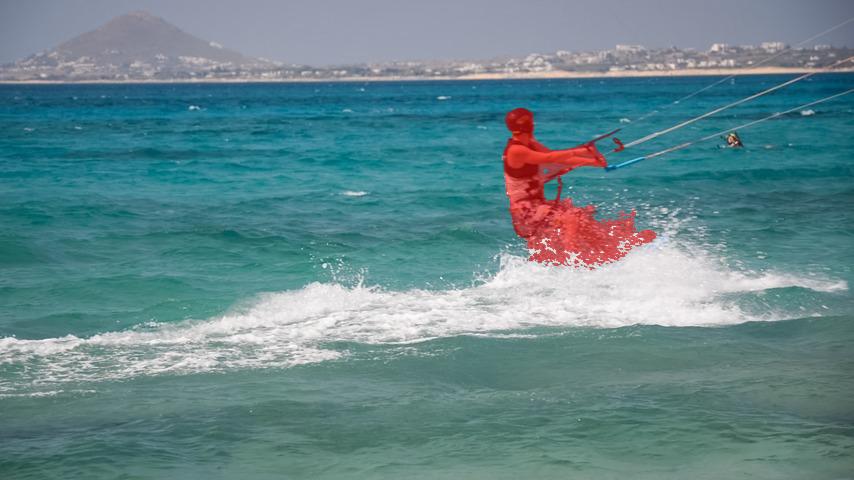} \\ 

\includegraphics[width=0.33\columnwidth]{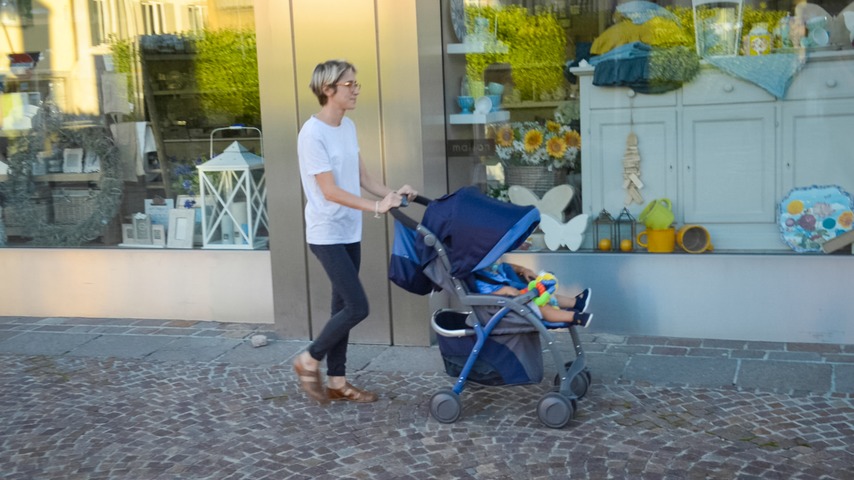} & 
\includegraphics[width=0.33\columnwidth]{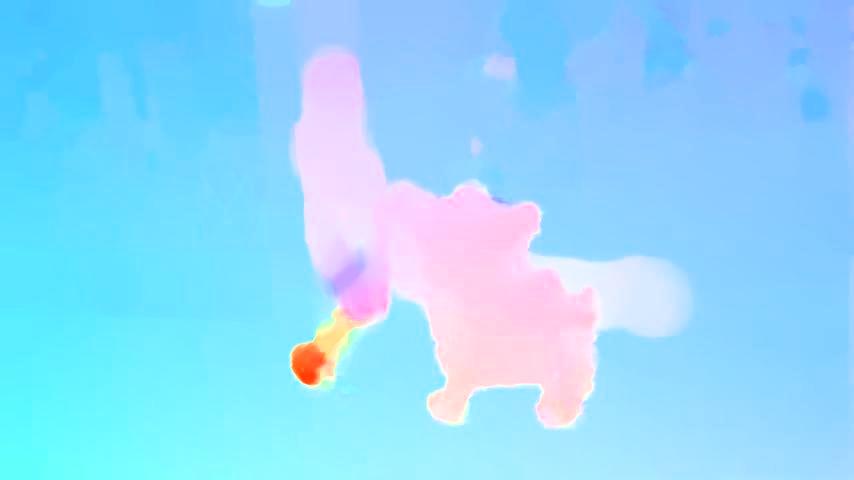} & 
\includegraphics[width=0.33\columnwidth]{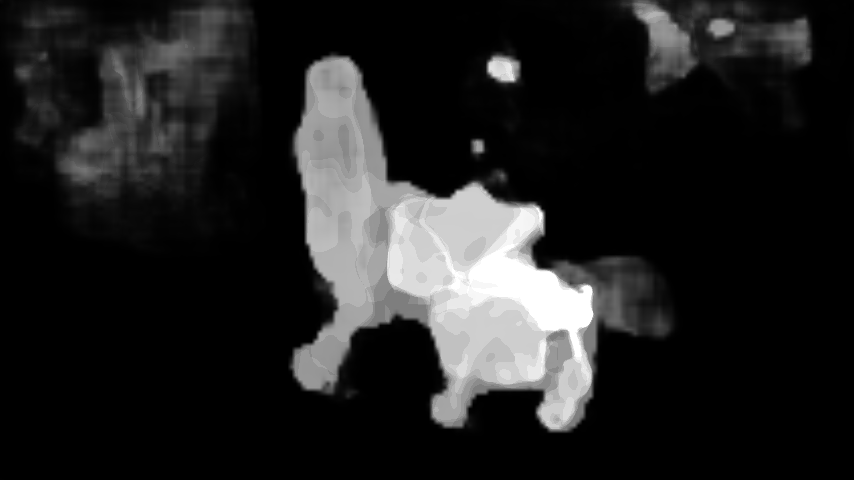} & 
\includegraphics[width=0.33\columnwidth]{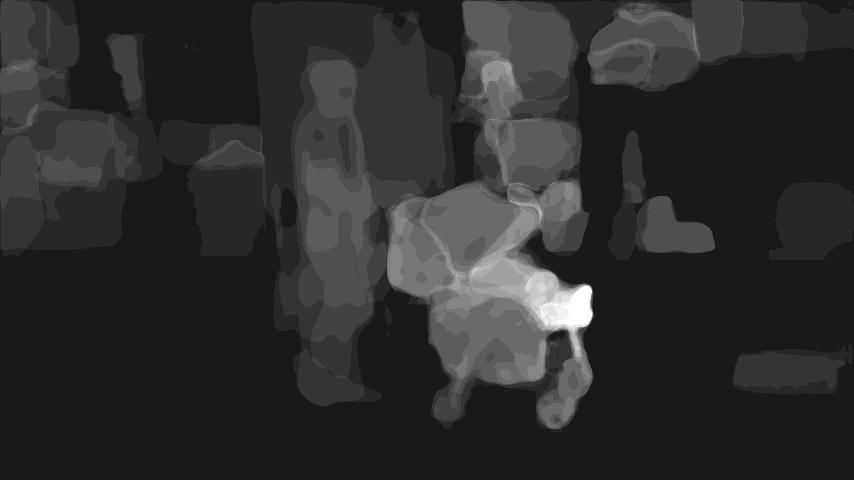} & 
\includegraphics[width=0.33\columnwidth]{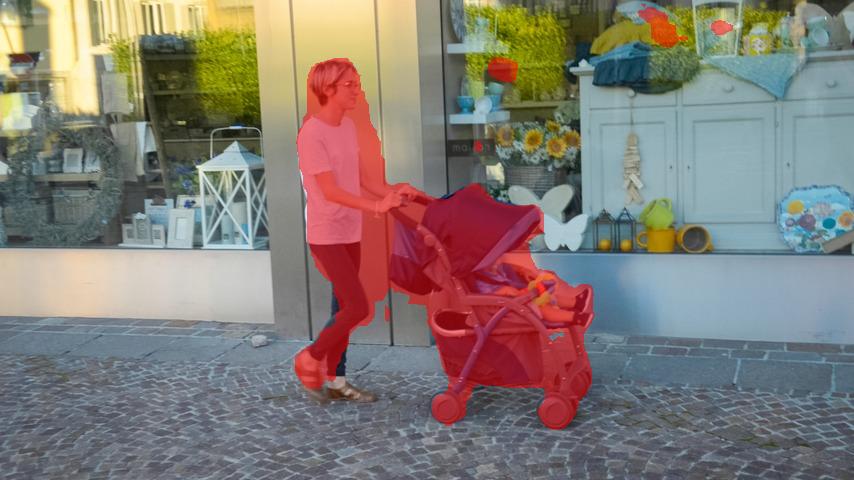} & 
\includegraphics[width=0.33\columnwidth]{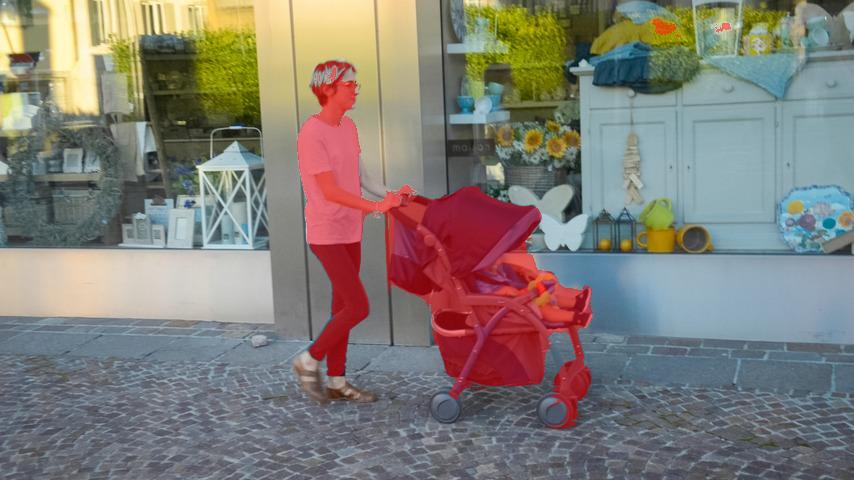} \\ 

(a) & (b) & (c) & (d) & (e) & (f)
\end{tabular}
\end{center}
\vspace{-0.3cm}\caption{Sample results on the DAVIS dataset showing all the
components of our approach. Each row shows: (a) video frame, (b) optical flow
estimated with LDOF~\cite{Brox11a}, (c) output of our MP-Net with LDOF
flow as input, (d) objectness
map computed with proposals~\cite{pinheiro2016learning}, (e) initial moving
object segmentation result, (f) segmentation refined with CRF.\vspace{-0.5cm}}
\label{fig:obj}
\end{figure*}
\setlength{\tabcolsep}{6pt}

\section{Detecting Motion Patterns}
\label{sec:realvid}
\vspace{-0.2cm}
We apply our trained model on synthetic (FlyingThings3D) as well as real-world
(DAVIS, BMS-26, FBMS) test data. Figure~\ref{fig:qualff} shows sample
predictions of our model on the FlyingThings3D test set with ground-truth
optical flow as input. Examples in the first two rows show that our model
accurately identifies fine details in objects: thin structures even when they
move subtlely, such as the neck of the guitar in the top-right corner in the
first row (see the subtle motion in the optical flow field (b)), fine
structures like leaves in the vase, and the guitar's headstock in the second
row.  Furthermore, our method successfully handles objects exhibiting highly
varying motions in the second example. The third row shows a limiting case,
where the receptive field of our network falls entirely within the interior of
a large object, as the moving object dominates. Traditional approaches, such as
RANSAC, do not work in this case either.

In order to detect motion patterns in real-world videos, we first compute
optical flow with popular methods~\cite{Brox11a,sundaram2010dense,Revaud15}.
With this flow as input to the network, we estimate a motion label map, as
shown in the examples in Figure~\ref{fig:obj}(c). Although the prediction of
our frame-pair feedforward model is accurate in several regions in the frame
((c) in the figure), we are faced with two challenges, which were not observed
in the synthetic training set. The first one is motion of
\textit{stuff}~\cite{Adelson01} in a scene, e.g., patterns on the water due to
the kiteboarder's motion (first row in the figure), which is irrelevant for
moving object segmentation. The second one is significant errors in optical
flow, e.g., in front of the pram ((b) in the bottom row in the figure). We
address these challenges by: (i) incorporating object
proposals~\cite{pinheiro2016learning} into our framework, and (ii) refining the
result with a fully-connected conditional random field
(CRF)~\cite{krahenbuhl2011efficient}. The following two sections present these
in detail, and their influence is analyzed in Section~\ref{sec:realvidexp}.

\vspace{-0.17cm}
\subsection{Segmenting real-world videos}
\label{sec:obj}
\vspace{-0.18cm}
As mentioned in the example above (Figure~\ref{fig:obj}, top row), real-world
videos may contain \textit{stuff} (water in this case) undergoing independent
motion. While it is interesting to study this motion, and indeed, our model
estimates it (see network prediction (c) in the first row), it is not annotated
in any of the standard datasets for moving object segmentation. In order to
perform a fair evaluation on standard benchmarks, we introduce the notion of
objects, with an objectness score, computed from object proposals, to eliminate
``moving stuff.'' We combine this score with our network output to obtain an
updated prediction.

We first generate object proposals in each frame with a state-of-the-art
method~\cite{pinheiro2016learning}. We then use a pixel-level voting scheme to
compute an objectness score. The score at a pixel $i$ is the number of
proposals that include it. This score is normalized by the total number of
proposals to obtain $o_i$, the objectness score at pixel $i$ in the $0-1$
range. In essence, we aggregate several proposals, which are likely to
represent objects of interest, to obtain an objectness map, as shown in the
examples in Figure~\ref{fig:obj}(d). We then combine this with the motion
prediction of our MP-Net at pixel $i$, $m_i \in [0, 1]$, to obtain an updated
prediction $p_i$ as: \mbox{$p_i = \min(m_i * (k + o_i), 1)$,} where $k \in [0,
1]$ is a parameter controlling the influence of objectness. It is set to $0.5$
to ensure that a high-confidence network prediction $m_i$ gets suppressed only
when there are no objects proposals supporting it. In the example with the
kiteboarder (top row in Figure~\ref{fig:obj}), the objectness map (d) has no
object proposals on water, the ``moving stuff,'' and eliminates it, to obtain
segmentation (e) that corresponds to the moving object.

\vspace{-0.05cm}
\subsection{Refining the segmentation}
\label{sec:crf}
\vspace{-0.15cm}
As shown on the synthetic test sequences, Figure~\ref{fig:qualff}, our model
produces accurate object boundaries in several cases. This is in part due to
precise optical flow input; recall that we use ground-truth flow for synthetic
data. Naturally, computed flow is less accurate than this, and can often fail
to provide precise object boundaries (see Figure~\ref{fig:obj}(b)),
especially in low-texture regions. Such errors inevitably result in imprecise
motion segments. To address this, we follow the common practice of refining
segmentation results with a CRF~\cite{chen2014semantic}. We use a
fully-connected CRF~\cite{krahenbuhl2011efficient}, with our predictions
updated with objectness scores as the unary terms, and standard colour-based
pairwise terms. The refinement is shown qualitatively in
Figure~\ref{fig:obj}(f), which improves over the initial segmentation in (e),
e.g., contours of the person pushing the pram in the middle row.

\section{Datasets}
\label{sec:datasets}
\vspace{-0.2cm}
\noindent {\bf FlyingThings3D (FT3D).} We train our network with the synthetic
FlyingThings3D dataset~\cite{Mayer16}. It contains videos of various objects
flying along randomized trajectories, in randomly constructed scenes. The video
sequences are generated with complex camera motion, which is also randomized.
FT3D comprises 2700 videos, each containing 10 stereo frames. The dataset is
split into training and test sets, with 2250 and 450 videos respectively.
Ground-truth optical flow, disparity, intrinsic and extrinsic camera
parameters, and object instance segmentation masks are provided for all the
videos. No annotation is directly available to distinguish moving objects from
stationary ones, which is required to train our network. We extract this from
the data provided as follows. With the given camera parameters and the stereo
image pair, we first compute the 3D coordinates of all the pixels in a video
frame $t$. Using ground-truth flow between frames $t$ and $t+1$ to find a pair
of corresponding pixels, we retrieve their respective 3D scene points. Now, if
the pixel has not undergone any independent motion between these two frames,
the scene coordinates will be identical (up to small rounding errors). We have
made these labels publicly available on our project website. Performance on the
test set is measured as the standard intersection over union score between the
predicted segmentation and the ground-truth masks.

\noindent {\bf DAVIS.} We use the densely annotated video segmentation
dataset~\cite{Perazzi16} exclusively for evaluating our approach. DAVIS is a
very recent dataset containing 50 full HD videos, featuring diverse types of
object and camera motion. It includes challenging examples with occlusion,
motion blur and appearance changes. Accurate pixel-level annotations are
provided for the moving object in all the video frames. Note that only a single
object is annotated in each video, even if there are multiple moving objects in
the scene. We evaluate our method on DAVIS with the three measures used
in~\cite{Perazzi16}, namely intersection over union for region similarity,
F-measure for contour accuracy, and temporal stability for measuring the
smoothness of segmentation over time. We follow the protocol
in~\cite{Perazzi16} and use images downsampled by a factor of two.

\noindent {\bf Other datasets.} We also evaluate on sequences from Berkeley
(BMS-26)~\cite{brox2010object,Tron07} and Freiburg-Berkeley
(FBMS)~\cite{ochs2014segmentation} motion segmentation datasets. The BMS-26
dataset consists of 26 videos with ground-truth object segmentations for a
selection of frames. Observing that annotations in some of these videos do not
correspond to objects with independent motion, ten of them were excluded
in~\cite{Bideau16}. In order to compare with~\cite{Bideau16}, we follow their
experimental protocol, and evaluate on the same subset of BMS-26. FBMS is an
extension of BMS-26, with 59 sequences in total, and a train and test split of
29 and 30 respectively. We use the test set in this paper. Performance on these
two datasets is evaluated with F-measure, as done
in~\cite{Bideau16,taylor2015causal}.

\begin{table}
\begin{center}
\begin{tabular}{c|l|c|c}
\hline
\# dec. & Trained on FT3D with ... & FT3D & DAVIS \\
\hline
\multirow{6}{*}{1} & RGB single frame  & 68.1 & 12.7  \\
 & RGB pair  & 69.1 & 16.6  \\
 & GT flow  & 74.5 & 44.3  \\
 & GT angle field  & 73.1 & 46.6  \\
 & RGB + GT angle field  & 74.8 & 39.6  \\
 & LDOF angle field  & 63.2 & 38.1  \\
\hline
4 & GT angle field  & 85.9 & 52.4  \\
\hline
\end{tabular}
\vspace{0.1cm}
\caption{Comparing the influence of different input modalities on the
FlyingThings3D (FT3D) test set and DAVIS. Performance is shown as mean
intersection over union scores. \# dec.\ refers to the number of decoder units
in our MP-Net. Ground-truth flow is used for evaluation on FT3D and LDOF flow
for DAVIS.}
\label{tbl:input}
\vspace{-0.85cm}
\end{center}
\end{table}

\vspace{-0.2cm}
\section{Experiments and Results}
\label{sec:expts}

\vspace{-0.16cm}
\subsection{Implementation details}
\label{sec:implement}
\vspace{-0.1cm}
\noindent {\bf Training.} We use mini-batch SGD with a batch size of 13
images---the maximum possible due to GPU memory constraints. The network is
trained from scratch with learning rate set to $0.003$, momentum to $0.9$, and
weight decay to $0.005$. Training is done for 27 epochs, and the learning rate
and weight decay are decreased by a factor of $0.1$ after every 9 epochs. We
downsample the original frames of the FT3D training set by a factor 2, and
perform data augmentation by random cropping and mirroring. Batch
normalization~\cite{ioffe2015batch} is applied to all the convolutional layers
of the network.

\vspace{0.06cm}
\noindent {\bf Other details.} We perform zero-mean normalization of the flow
field vectors, similar to~\cite{simonyan2014two}. When using flow angle and
magnitude together (which we refer to as flow angle field), we scale the
magnitude component, to bring the two channels to the same range. We use 100
proposals in each frame to compute the objectness score (see \S\ref{sec:obj}).
Also, for a fair comparison to other methods on DAVIS, we do not learn the
parameters of the fully-connected CRF on this dataset, and instead set them to
values used for a related pixel-level segmentation
task~\cite{chen2014semantic}. Our model is implemented in the Torch framework.

\vspace{-0.1cm}
\subsection{Influence of input modalities}
\label{sec:mod}
\vspace{-0.15cm}
We first analyze the influence of different input modalities on training our
network. Specifically, we use RGB data (single frame and image pair), optical
flow field (ground truth and estimated one), directly as flow vectors, i.e.,
flow in x and y axes, or as angle field (flow vector angle concatenated with
flow magnitude), and a combination of RGB data and flow. These results are
presented on the FT3D test set and
also on DAVIS, to study how well the observations on synthetic videos transfer
to the real-world ones, in Table~\ref{tbl:input}. For computational reasons we
train and test with different modalities on a smaller version of our MP-Net,
with one decoder unit instead of four. Then we pick the best modality to train
and test the full, deeper version of the network.

\begin{table}[t]
\begin{center}
\begin{tabular}{l|l|c}
\hline
Variant of our method & Flow used & Mean IoU \\
\hline
MP-Net & LDOF  & 52.4  \\
MP-Net & EpicFlow  & 56.9  \\
MP-Net + Objectness & LDOF  & 63.3 \\
MP-Net + Objectness & EpicFlow  & 64.5 \\
MP-Net + Objectness + CRF & LDOF  & 69.7  \\
MP-Net + Objectness + CRF & EpicFlow  & 68.0  \\
\hline
\end{tabular}
\vspace{0.1cm}
\caption{Performance of our best network (4 decoder units trained on GT angle
field) with additional cues (Objectness, CRF) and different flow inputs (LDOF,
EpicFlow) on DAVIS.\vspace{-1.00cm}}
\label{tbl:davis}
\end{center}
\end{table}

\begin{table*}
\begin{center}
\begin{tabular}{l | c | c cccccccc}
\hline
\multicolumn{2}{c|}{Measure} & NLC~\cite{Faktor14}  & CVOS~\cite{taylor2015causal} & TRC~\cite{fragkiadaki2012video} & MSG~\cite{brox2010object} & KEY~\cite{lee2011key} & SAL~\cite{wang2015saliency} & FST~\cite{papazoglou2013fast} & PCM~\cite{Bideau16} & Ours \\
\hline
\multirow{3}{*}{$\mathcal{J}$} & Mean & 64.1 & 51.4 & 50.1 & 54.3 & 56.9 & 42.6 & 57.5 & 45.5 & \textbf{69.7}  \\
& Recall & 73.1 & 58.1 & 56.0 & 63.6 & 67.1 & 38.6 & 65.2 & 44.3 & \textbf{82.9}  \\
& Decay & ~8.6 & 12.7 & ~~5.0 & ~\textbf{2.8} & ~7.5 & ~8.4 & ~4.4 & 11.8 & ~5.6  \\
\hline
\multirow{3}{*}{$\mathcal{F}$} & Mean & 59.3 & 49.0 & 47.8 & 52.5 & 50.3 & 38.3 & 53.6 & 46.1 & \textbf{66.3}  \\
& Recall & 65.8 & 57.8 & 51.9 & 61.3 & 53.4 & 26.4 & 57.9 & 43.7 & \textbf{78.3}  \\
& Decay & ~8.6 & 13.8 & ~~6.6 & ~\textbf{5.7} & ~7.9 & ~7.2 & ~6.5 & 10.7 & ~6.7  \\
\hline
$\mathcal{T}$ & Mean & 35.6 & 24.3 & 32.7 & 25.0 & \textbf{19.0} & 60.0 & 27.6 & 51.3 & 68.6  \\
\hline
\end{tabular}
\caption{Comparison to state-of-the-art methods on DAVIS with intersection over
union ($\mathcal{J}$), F-measure ($\mathcal{F}$), and temporal stability
($\mathcal{T}$).}
\label{tbl:soadavis}
\vspace{-0.7cm}
\end{center}
\end{table*}
\begin{figure*}[t]
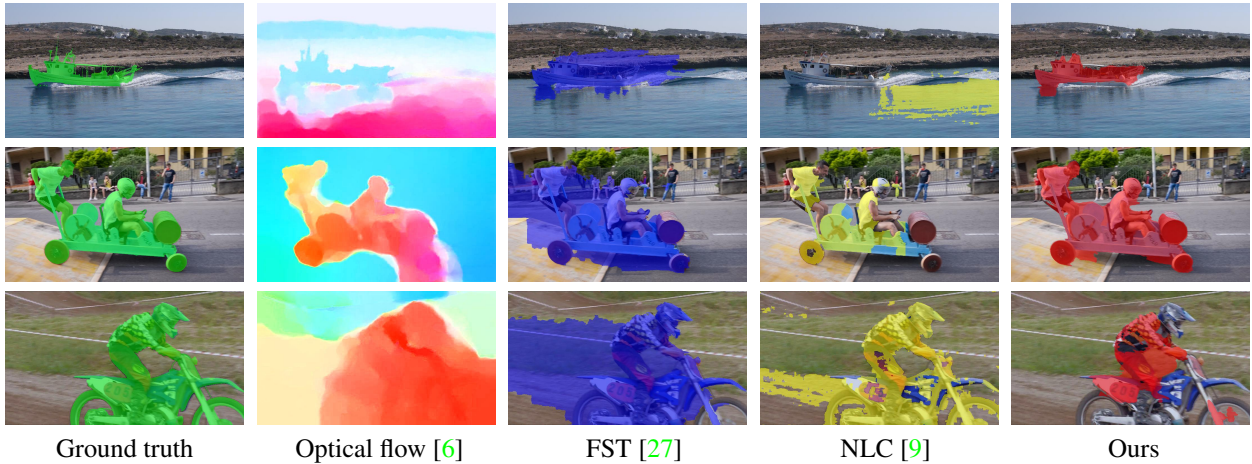

\begin{center}
\fivefigures{soa/boat/gt}{soa/boat/flow}{soa/boat/vito}{soa/boat/irani}{soa/boat/ours}{0.38\columnwidth}\vspace{0.1cm}
\fivefigures{soa/soapbox/gt}{soa/soapbox/flow}{soa/soapbox/vito}{soa/soapbox/irani}{soa/soapbox/ours}{0.38\columnwidth}\vspace{0.1cm}
\fivefigurescaptionDavis{soa/motocross-bumps/gt}{soa/motocross-bumps/flow}{soa/motocross-bumps/vito}{soa/motocross-bumps/irani}{soa/motocross-bumps/ours}{0.38\columnwidth}
\end{center}
\vspace{-0.3cm}\caption{Qualitative comparison with top-performing methods on
DAVIS. Left to right: ground truth, optical flow~\cite{Brox11a}, results of
FST~\cite{papazoglou2013fast}, NLC~\cite{Faktor14}, and our approach. The last
row shows a failure case of our approach, i.e., a part of the motorbike is
missing.}
\vspace{-0.5cm}
\label{fig:davis}
\end{figure*}

\begin{table*}
\begin{center}
\begin{tabular}{c|c c c c c c c c}
\hline
Measure & CUT~\cite{keuper2015motion}  & FST~\cite{papazoglou2013fast} & TRC~\cite{fragkiadaki2012video} & MTM~\cite{zamalieva2014multi} & CMS~\cite{Narayana13} & PCM~\cite{Bideau16} & MP+Obj & MP+Obj + FST~\cite{papazoglou2013fast} \\
\hline
$\mathcal{F}$ & 73.0 & 64.1 & 72.8 & 66.0 & 62.5 & \textbf{78.2} & 71.8 & \textbf{78.1} \\
\hline
\end{tabular}
\vspace{0.1cm}
\caption{Comparison to state-of-the-art methods on the subset of BMS-26 used
in~\cite{Bideau16} with F-measure. `MP+Obj' is MP-Net with objectness.}
\label{tbl:bms}
\vspace{-0.7cm}
\end{center}
\end{table*}
\begin{figure*}[t]
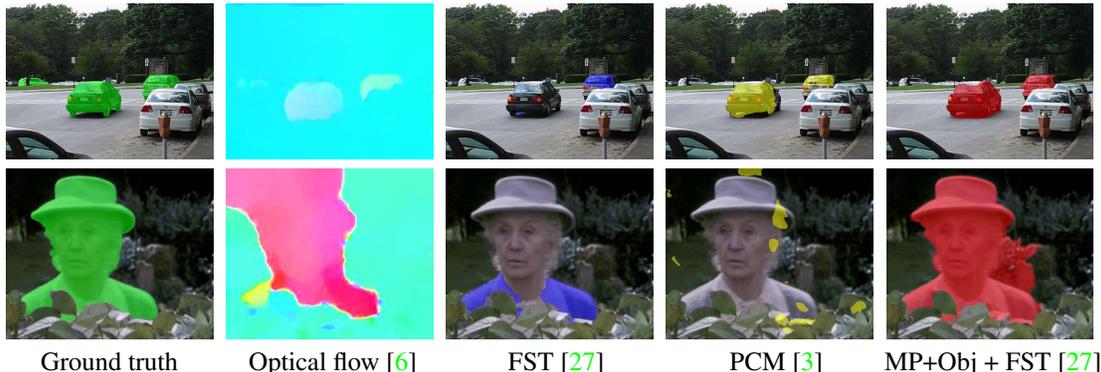

\begin{center}
\fivefigures{bms_soa/cars2/gt}{bms_soa/cars2/flow}{bms_soa/cars2/vito}{bms_soa/cars2/pia}{bms_soa/cars2/ours}{0.33\columnwidth}\vspace{0.1cm}
\fivefigurescaption{bms_soa/marple4/gt}{bms_soa/marple4/flow}{bms_soa/marple4/vito}{bms_soa/marple4/pia}{bms_soa/marple4/ours}{0.33\columnwidth}
\end{center}
\vspace{-0.3cm}\caption{Qualitative comparison on two sample sequences from
BMS-26. Left to right: ground truth, optical flow~\cite{Brox11a}, results of
FST~\cite{papazoglou2013fast}, PCM~\cite{Bideau16}, and our MP-Net + Objectness
+ FST (`MP+Obj + FST~\cite{papazoglou2013fast}').}
\vspace{-0.5cm}
\label{fig:bms}
\end{figure*}

From Table~\ref{tbl:input}, the performance on DAVIS is lower than on FT3D.
This is expected as there is domain change from synthetic to real data, and
that we use ground truth optical flow as input for FT3D test data, but
estimated flow~\cite{Brox11a,sundaram2010dense} for DAVIS. As a baseline, we
train on single RGB frames (`RGB single frame' in the table). Clearly, no
motion patterns can be learned in this case, but the network performs
reasonably on FT3D test (68.1), as it learns to correlate object appearance
with its motion. This intuition is confirmed by the fact that `RGB single
frame' fails on DAVIS (12.7), where the appearance of objects and background is
significantly different from FT3D.  MP-Net trained on `RGB pair', i.e., RGB
data of two consecutive frames concatenated, performs slightly better on both
FT3D (69.1) and DAVIS (16.6), suggesting that it captures some motion-like
information, but continues to rely on appearance, as it does not transfer well
to DAVIS.

Training on ground-truth flow vectors corresponding to the image pair (`GT
flow') improves the performance on FT3D by 5.4\% and on DAVIS significantly
(27.7\%). This shows that MP-Net learned on flow from synthetic examples can be
transferred to real-world videos. We then experiment with flow angle as part of
the input. As discussed in~\cite{Narayana13}, flow orientations are independent
of depth from the camera, unlike flow vectors, when the camera is undergoing
only translational motion. Using the ground truth flow angle field (concatenation
of flow angles and magnitudes) as input (`GT angle field'), we note a slight
decrease in IoU score on FT3D (1.4\%), where strong camera rotations are
abundant, but in real examples, such motion is usually mild. Hence, `GT angle
field' improves IoU on DAVIS by 2.3\%. We use angle field representation in all
further experiments.

Using a concatenated flow and RGB representation (`RGB + GT angle field')
performs better on FT3D (by 1.7\%), but is poorer by 7\% on DAVIS,
re-confirming our observation that appearance features are not consistent
between the two datasets. Finally, training on computed flow~\cite{Brox11a}
(`LDOF angle field') leads to significant drop on both the datasets: 9.9\% on
FT3D (with GT flow for testing) and 8.5\% on DAVIS, showing the
importance of high-quality training data for learning accurate
models. The full version of our MP-Net, with 4 decoder units, improves
the IoU by 12.8\% on FT3D and 5.8\% on DAVIS over its shallower
one-unit equivalent.  

Notice that the performance of our full model on FT3D is excellent, with the
remaining errors mostly due to inherently ambiguous cases like objects moving
close to the camera (see third row in Figure~\ref{fig:qualff}), or very strong
object/camera motion. On DAVIS the results are considerably lower despite less
challenging motion. To investigate the extent to which this is due to errors
in flow estimation, we compute LDOF~\cite{Brox11a} flow on the FT3D test set
and evaluate our full model trained on ground-truth flow. We observe a
significant drop in performance by 27.2\% (from 85.9\% to 58.7\%). This
confirms the impact of optical flow quality and suggests that improvements in
flow estimation can increase the performance of our method on real-world
videos.

\vspace{-0.2cm}
\subsection{Evaluation on real videos}
\label{sec:realvidexp}
\vspace{-0.2cm}
We show the performance of our MP-Net on DAVIS in Table~\ref{tbl:davis}, along
with a study on the influence of additional cues and the flow used. First, we
evaluate the importance of the estimated flow quality by comparing
EpicFlow~\cite{Revaud15}, a recent method, and
LDOF~\cite{Brox11a,sundaram2010dense}, a more classical method. Using EpicFlow,
which leverages motion contours, produces more accurate object boundaries, and
improves over MP-Net using LDOF by 4.5\%. Incorporating objectness cues with
our network (`MP-Net + Objectness' in the table), as described in
Section~\ref{sec:obj} improves the segmentation results over `MP-Net' by 10.9\%
and 7.6\% with LDOF and EpicFlow respectively. Refining these segmentation
results with a fully-connected CRF (`MP-Net + Objectness + CRF'), as in
Section~\ref{sec:crf}, further improves the IoU by 6.4\% and 3.5\% with LDOF
and EpicFlow respectively. This refinement has a significant impact when using
LDOF flow, as it improves segmentation around object boundaries, as shown in
Figure~\ref{fig:obj}. On the other hand, EpicFlow already incorporates motion
boundaries, and a CRF refinement on top of the results with this flow has a
less-pronounced improvement. The overall method `MP-Net + Objectness + CRF'
performs better with LDOF (69.7) than EpicFlow (68.0). Although EpicFlow has
better precision than LDOF around object boundaries, it tends to make larger
errors in other regions, which cannot be corrected with CRF refinement.  We
thus use LDOF in the following experiments.

\vspace{-0.2cm}
\subsection{Comparison to the state of the art}
\label{sec:soa}
\vspace{-0.2cm}
Table~\ref{tbl:soadavis} shows comparison with unsupervised state-of-the-art
methods on DAVIS. In addition to comparing with the methods reported
in~\cite{Perazzi16}, we evaluate PCM~\cite{Bideau16}, the top-performer on
BMS-26, with source code provided by the authors. Note that methods which use
supervision on DAVIS test sequences (e.g., annotation in the first frame) do
perform better, but are not directly comparable to our method. Our frame-level
approach `MP-Net + Objectness + CRF' (Ours) outperforms all the methods
significantly, notably by 5.6\% on IoU (mean-$\mathcal{J}$) and 7\% on
F-measure (mean-$\mathcal{F}$) over the best one in the
evaluation~\cite{Perazzi16}, i.e., NLC~\cite{Faktor14}. Note that the two top
methods, NLC and FST~\cite{papazoglou2013fast} perform a video-level inference
by propagating motion labels through the video, unlike our approach using only
a pair of video frames at a time. Our network shows the top performance, by a
significant margin, with respect to mean and recall on the IoU and F-measure
scores. All the methods perform similarly on the decay scores, which quantifies
the performance loss/gain over time. As MP-Net uses limited temporal
information (two-frame optical flow) and does not perform inference at the
video level, it is not the best one on the temporal stability measure. This
limitation can be addressed with a post-processing step, such as using a
temporal CRF.

Figure~\ref{fig:davis} compares our approach qualitatively to the two
top-performing methods, FST~\cite{papazoglou2013fast} and NLC~\cite{Faktor14},
on DAVIS. In the first row, FST localizes the moving boat, but its segmentation
leaks into the background region around the boat, due to errors in motion
tracking. NLC latches onto moving water, whereas our MP-Net segments the boat
accurately.
In the second row, our segmentation result is more precise and complete than
both FST and NLC. The last row shows a failure case, where a part of the
motorbike is missing due to highly imprecise flow estimation.

Table~\ref{tbl:bms} shows quantitative comparison on the subset of BMS-26 used
in~\cite{Bideau16}. We observed that objects are annotated in some of the
sequences when they do not undergo independent motion. Thus, the results of our
MP-Net with objectness (`MP+Obj' in the table) are not directly comparable to
the other methods which use propagation between frames, though they are still
on par with many of the previous methods.  To account for this mismatch with
MP-Net, which only segments moving objects, we incorporate our frame-level
motion estimation results into a state-of-the-art video segmentation
method~\cite{papazoglou2013fast}. This is achieved by replacing the location
unary scores in~\cite{papazoglou2013fast} with our motion prediction scores
integrated with objectness. The results (`MP+Obj +
FST~\cite{papazoglou2013fast}' in the table) are significantly better than most
previous methods, and on par with PCM~\cite{Bideau16}. In particular, using our
motion prediction in~\cite{papazoglou2013fast} improves the result by 14\%. We
also evaluated this combination (`MP+Obj + FST') on the FBMS test set, where it
achieves 77.5\% in F-measure, and is better than state-of-the-art methods:
FST~\cite{papazoglou2013fast} (69.2), CVOS~\cite{taylor2015causal} (74.9), and
CUT~\cite{keuper2015motion} (76.8).

Figure~\ref{fig:bms} compares our results on BMS-26 with the top-method on this
dataset, PCM~\cite{Bideau16}, and the baseline video-level approach
FST~\cite{papazoglou2013fast}. In the first row, FST segments only one of the
two moving cars in the foreground, due to very slow motion of the second car.
Introducing our motion prediction into FST segments both these cars. This
result is comparable to PCM. None of the methods segments the third car in the
background however. In the second row, PCM fails to segment the woman, and FST
segments only the jacket, but including our motion estimate into FST
significantly improves the result. Tracking errors inherent in FST result in
the segmentation leaking into the background.

\vspace{-0.2cm}
\section{Conclusion}
\vspace{-0.2cm}
This paper introduces a novel approach for learning motion patterns in videos.
Its strength is demonstrated for the task of moving object segmentation, where
our method outperforms many complex approaches, that rely on engineered
features. Future work includes: (i)~development of end-to-end trainable models
for video semantic segmentation, (ii)~use of a memory module for video object
segmentation, (iii)~using additional information, e.g., subset of frames
annotated by users, to handle ambiguous cases.

\vspace{0.2cm}
{\small \noindent {\bf Acknowledgments.}
This work was supported in part by the ERC advanced grant ALLEGRO, the
MSR-Inria joint project, a Google research award and a Facebook gift. We
gratefully acknowledge the support of NVIDIA with the donation of GPUs used for
this research.}

{\small
\bibliographystyle{ieee}
\bibliography{motionest}
}

\end{document}

%% file: macros.tex
\newcommand{\twofigures}[3]{
            \centerline{{\includegraphics[width=#3]{#1}}~~{\includegraphics[width=#3]{#2}}}}

\newcommand{\threefiguresh}[4]{
            \centerline{{\includegraphics[height=#4]{#1}}{\includegraphics[height=#4]{#2}}{\includegraphics[height=#4]{#3}}}}

\newcommand{\threefigures}[4]{
            \centerline{{\includegraphics[width=#4]{#1}}~~{\includegraphics[width=#4]{#2}}~~{\includegraphics[width=#4]{#3}}}}

\newcommand{\threefigureslbl}[5]{
            \centerline{(#5){\includegraphics[width=#4]{#1}}~~{\includegraphics[width=#4]{#2}}~~{\includegraphics[width=#4]{#3}}}}

\newcommand{\lthreefigures}[4]{
            \centerline{{\includegraphics[width=#4]{#1}}{\includegraphics[width=#4]{#2}}{\includegraphics[width=#4]{#3}}
			\makebox[0.33\columnwidth][c]{(a)}\makebox[0.33\columnwidth][c]{(b)}\makebox[0.33\columnwidth][c]{(c)}}}

\newcommand{\fourfigures}[5]{
            \centerline{{\includegraphics[width=#5]{#1}}~~{\includegraphics[width=#5]{#2}}~~{\includegraphics[width=#5]{#3}}~~{\includegraphics[width=#5]{#4}}}}

\newcommand{\fourfigurescaption}[9]{
            \centerline{{\includegraphics[width=#5]{#1}}~~{\includegraphics[width=#5]{#2}}~~{\includegraphics[width=#5]{#3}}~~{\includegraphics[width=#5]{#4}}}
            \makebox[#5][c]{#6}\makebox[#5][c]{#7}\makebox[#5][c]{#8}\makebox[#5][c]{#9}}
            
\newcommand{\fivefigures}[6]{
            \centerline{{\includegraphics[width=#6]{#1}}~~{\includegraphics[width=#6]{#2}}~~{\includegraphics[width=#6]{#3}}~~{\includegraphics[width=#6]{#4}}~~{\includegraphics[width=#6]{#5}}}}      
  
\newcommand{\fivefigurescaptionDavis}[6]{
            \centerline{{\includegraphics[width=#6]{#1}}~~{\includegraphics[width=#6]{#2}}~~{\includegraphics[width=#6]{#3}}~~{\includegraphics[width=#6]{#4}}~~{\includegraphics[width=#6]{#5}}}\makebox[#6][c]{Ground truth~~~~~~~~}\makebox[#6][c]{Optical flow~\cite{Brox11a}~~~}\makebox[#6][c]{~~FST~\cite{papazoglou2013fast}}\makebox[#6][c]{~~~~~~NLC~\cite{Faktor14}}\makebox[#6][c]{~~~~~~~~~Ours}}
            
\newcommand{\fivefigurescaption}[6]{
            \centerline{{\includegraphics[width=#6]{#1}}~~{\includegraphics[width=#6]{#2}}~~{\includegraphics[width=#6]{#3}}~~{\includegraphics[width=#6]{#4}}~~{\includegraphics[width=#6]{#5}}}\makebox[#6][c]{Ground truth~~~~~~~~}\makebox[#6][c]{Optical flow~\cite{Brox11a}~~~}\makebox[#6][c]{~~FST~\cite{papazoglou2013fast}}\makebox[#6][c]{~~~~~~PCM~\cite{Bideau16}}\makebox[#6][c]{~~~~~~~~~MP+Obj + FST~\cite{papazoglou2013fast}}}

\newcommand{\sixfigures}[7]{
            \centerline{{\includegraphics[width=#7]{#1}}~~{\includegraphics[width=#7]{#2}}~~{\includegraphics[width=#7]{#3}}~~{\includegraphics[width=#7]{#4}}~~{\includegraphics[width=#7]{#5}}~~{\includegraphics[width=#7]{#6}}}}
            
\newcommand{\sixfiguresdots}[7]{
            \centerline{{\includegraphics[width=#7]{#1}}~~{\includegraphics[width=#7]{#2}}~~{\includegraphics[width=#7]{#3}}~~{...}~~{\includegraphics[width=#7]{#4}}~~{\includegraphics[width=#7]{#5}}~~{\includegraphics[width=#7]{#6}}}}

\newcommand{\threefigurescaption}[7]{
            \centerline{{\includegraphics[width=#4]{#1}}~~{\includegraphics[width=#4]{#2}}~~{\includegraphics[width=#4]{#3}}}
		     \makebox[#4][c]{#5}\makebox[#4][c]{#6}\makebox[#4][c]{#7}}

\newcommand{\comment}[1]{}

\newcommand{\twofigurescaption}[6]{
            \centerline{{\includegraphics[width=#3]{#1}}~~{\includegraphics[width=#3]{#2}}}
            \makebox[#6][c]{#4}\makebox[#6][c]{#5}}

\newcommand{\todo}[1]{{\bf TODO:} #1}